\documentclass[review]{elsarticle}

\usepackage{graphicx}
\usepackage{caption}
\usepackage{subcaption}
\usepackage{hyperref}
\usepackage{epstopdf}
\usepackage{url}
\usepackage{float}
\usepackage{amsfonts}

\begin{document}

\begin{frontmatter}

\title{Semi-Supervised Recognition of the Diploglossus Millepunctatus Lizard Species using Artificial Vision Algorithms}

\author[affiliation1]{Jhony-Heriberto Giraldo-Zuluaga \corref{mycorrespondingauthor}}
\cortext[mycorrespondingauthor]{Corresponding author}
\ead{jhonygiraldoz@gmail.com, heriberto.giraldo@udea.edu.co}

\author[affiliation1]{Augusto Salazar}

\author[affiliation2]{Juan M. Daza}

\address[affiliation1]{Grupo de Investigaci\'on SISTEMIC, Facultad de Ingenier\'ia, Universidad de Antioquia UdeA, Calle 70 No. 52-21, Medell\'in, Colombia}
\address[affiliation2]{Grupo Herpetol\'ogico de Antioquia, Instituto de Biolog\'ia, Universidad de Antioquia, Calle 67 No. 53-108, Bloque 7-121, A.A. 1226, Medell\'in, Colombia}

\begin{abstract}
Animal biometrics is an important requirement for monitoring and conservation tasks. The classical animal biometrics risk the animals' integrity, are expensive for numerous animals, and depend on expert criterion. The non-invasive biometrics techniques offer alternatives to manage the aforementioned problems. In this paper we propose an automatic segmentation and identification algorithm based on artificial vision algorithms to recognize \textit{Diploglossus millepunctatus}. \textit{Diploglossus millepunctatus} is an endangered lizard species. The algorithm is based on two stages: automatic segmentation to remove the subjective evaluation, and one identification stage to reduce the analysis time. A 82.87\% of correct segmentation in average is reached. Meanwhile the identification algorithm is achieved with euclidean distance point algorithms such as Iterative Closest Point and Procrustes Analysis. A performance of 92.99\% on the top 1, and a 96.82\% on the top 5 is reached. The developed software, and the database used in this paper are publicly available for download from the web page of the project.

\end{abstract}

\begin{keyword}
Animal biometrics \sep Diploglossus Millepunctatus \sep Active Contours \sep spots segmentation \sep registration algorithms \sep Iterative Closest Point \sep Procrustes Analysis
\end{keyword}

\end{frontmatter}

\section{Introduction}

The classical biometric identification systems are grouped on permanent methods, temporary methods, and electrical methods. The common problems with these methods stem from their vulnerability to losses, deformations, and fraud, not to mention animal-welfare concerns \cite{awad2016classical}.

Biometric methods of animal identification with artificial vision algorithms are application dependent, because each animal species has unique characteristics such as spots, stripes, lines or textures. It can search for characteristics to distinguish one individual from another one in the same species. In the literature, there are some examples, e.g. multi-curve matching of the elephants ears is used for identifying elephants \citep{ardovini2008identifying}. One approach to build a general animal biometrics system is to use the skin or fur patterns. Murray explained the structure of some animal patterns by hypothetical Turing systems named as coat patterns \citep{murray1988leopard}. 
The coat patterns are easy to acquire, and represent the individuality of an animal with respect to another of the same species. Examples of this are: monitoring populations of penguins using their coat patterns \citep{sherley2010spotting,burghardt2007individual}, recognition of whale sharks using the spot patterns of their skin \citep{arzoumanian2005astronomical}, among others. Murray explains that coat pattern formation is modeled as a reaction-diffusion mechanism that depends on some embryonic parameters. This phenomena can be amplified or inhibited forming the specific coat pattern. Animal biometrics is an emerging research field that combines pattern recognition, ecology and information sciences \citep{kuhl2013animal}.

The biometric recognition of lizards is sparsely reported in the literature. One example is the Pygmy bluetongue lizard, this lizard was identified by signature curves in \citep{li2009non}. The dotted lizard (\textit{Diploglossus millepunctatus}) is a lizard species native to Malpelo Island, located in the Colombian Pacific. The identification of this animal species is important, because it is critically endangered due to human activities \citep{lopez2006lizards}. To our knowledge, there is not a non-invasive biometric identification method in the literature to distinguish individuals of \textit{Diploglossus millepunctatus}. The method proposed here is non-invasive and based on image processing. Our algorithm includes segmentation and classification steps to identify each lizard. Various problems on spot identification can occur due to lighting variation \citep{de2010computer} or problems due to perspective \citep{kelly2001computer} that are discussed in this paper. The idea is to use segment spot patterns of \textit{Diploglossus millepunctatus} skin to solve the named problem (biometric recognition). In our method, a gamma correction algorithm was applied to get rid of lighting overexposure. Problems due to perspective were addressed by a deformable active contours model. For the identification step registration algorithms were used to align the spot clouds.

Some algorithms avoid the segmentation part using computer aided procedures \citep{de2010computer,kelly2001computer,bolger2012computer}. This paper proposes an automatic algorithm that approaches the segmentation problem by dividing the segmentation into two threads shown in Figure \ref{fig:segmentationAlgorithm}. These two threads have the aim of extract the spots on the lightest regions (upper thread) and darkest regions (lower thread). Each thread consists of three stages. One pre-processing procedure prepares the image for the active contours iteration; the pre-processing  procedure is the first stage. The principal Active Contours algorithm segments the image. The Active Contours are not trained but their parameters are tuned by an optimization algorithm; the Active Contours are the second stage. The third stage consists of an area opening procedure that deletes atypical regions \cite{zuluaga2016automatic}.

\begin{figure}[h]
\centering
\includegraphics[width=0.5\textwidth]{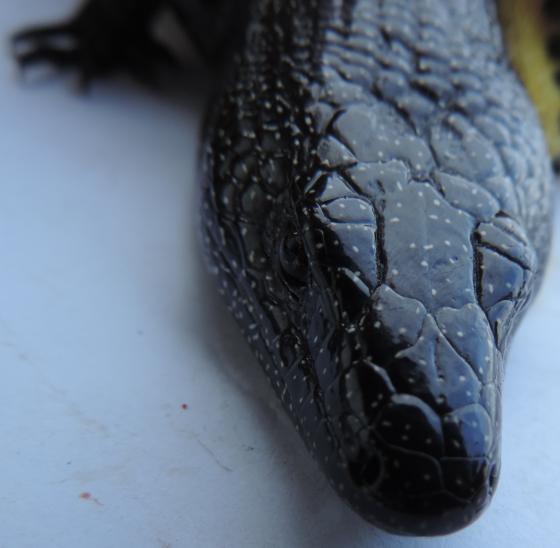}
\caption{Diploglossus millepunctatus}
\label{fig:diploMille}
\end{figure}

Spots segmentation has been used for medical purposes. One of the most important applications is disease diagnosis \citep{bell2007segmentation}. For disease diagnosis some methods are found in the literature. Borovec et al. use Bayesian classification combined with Markov Random Field \citep{borovec2013fully}. Veta et al. uses a watershed segmentation algorithm \citep{veta2011marker} and Mouelhi et al. an Active Contours algorithm \citep{mouelhi2013automatic}. The principal target of each algorithm is similar to our proposed method, but those algorithms have been used for medical purposes. The algorithm presented in this paper is used for biological purposes. Our images have illumination problems and the image is deformed by the perspective, thus the results of the segmentation of images for medical purposes and our algorithm cannot be numerically comparable. Figure \ref{fig:diploMille} shows \textit{Diploglossus millepunctatus}. As can be seen, the lizard’s skin is covered by a naturally oily substance that gives it an unavoidable shine. The idea of this paper is to segment the spots on the frontal scales and achieve the identification with those spot-patterns. The lizard’s scales represent sets of spots that are limited by grooves. The frontal scale is shown in Figure \ref{fig:frontalScale}.

This paper proposes two automatic identification algorithms. One identification algorithm is a matching based on Iterative Closest Point (ICP) \cite{best1992method}, and the other one is an algorithm based on ICP and Procrustes analysis \cite{dryden1998statistical}. There are other important stages in the identification algorithm such as normalization with respect to the database, and centroid point extraction. The registration algorithms, such as ICP and Procrustes analysis, have been used on reconstruction tasks in two and three dimensions for robot localization, for co-registering bone models, for computer aided tomography scans, etc. The ICP algorithm has been used in biometric tasks such as the fingerprint recognition \citep{jain2007pores}, human ears recognition in three dimensions \citep{chen2007human}, face recognition in two and three dimensions \citep{mian2007efficient}, among others. The source codes, the \textit{MilPuntos} software that compiles the algorithms, and the database used in this paper are publicly available for download from the web page of the project \cite{Salazar2016lizardDatabase}.

The paper is organized as follows. Section 2 shows material and methods. Section 3 describes the experimental framework. Section 4 presents the experimental results and the discussion. Finally, Section 5 shows the conclusion and future works.

\section{Material and methods}

\begin{figure}[h]
\centering
\includegraphics[width=0.7\textwidth]{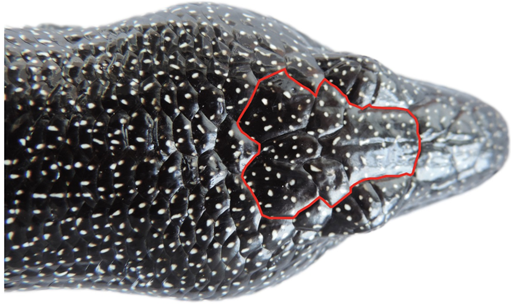}
\caption{Frontal scale.}
\label{fig:frontalScale}
\end{figure}

\begin{figure}[h]
    \centering
    \includegraphics[scale=0.7]{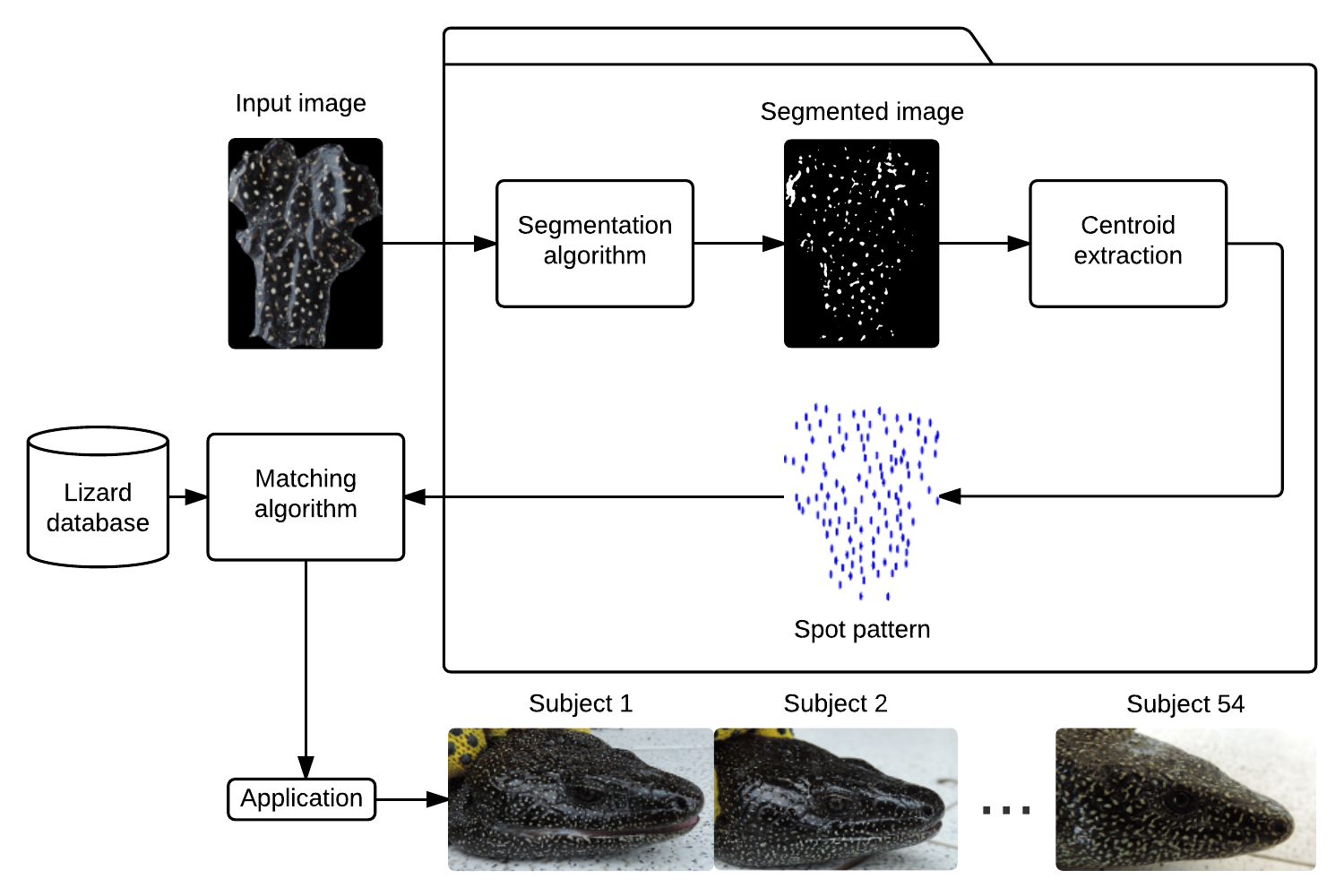}
    \caption{General diagram.}
    \label{fig:generalDiagram}
\end{figure}

The identification of lizards is based on the automatically extracted spots from the frontal scales shown in Figure \ref{fig:frontalScale}. An automatic spots extraction algorithm was designed to minimize the human intervention. The spots patterns of the lizard skin are used to match and identify the individual. Figure \ref{fig:generalDiagram} shows the general scheme of the main algorithm. This section explains the methods used on each process stage in this paper.

\subsection{Segmentation algorithm}\label{sec:segmAlgor}

\begin{figure}
    \centering
    \begin{subfigure}[b]{0.3\textwidth}
		\includegraphics[width=\textwidth]{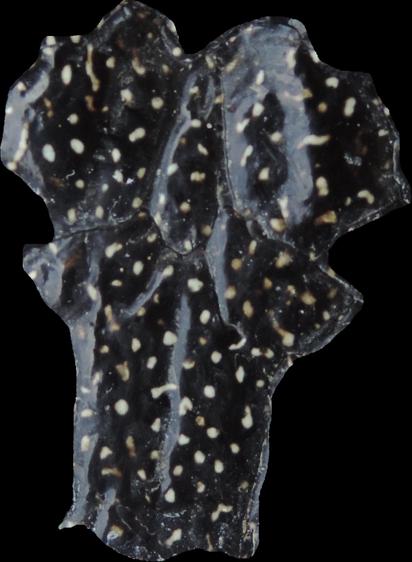}
        \caption{•}
        \label{fig:OriginalImage}
    \end{subfigure}
    \begin{subfigure}[b]{0.3\textwidth}
		\includegraphics[width=\textwidth]{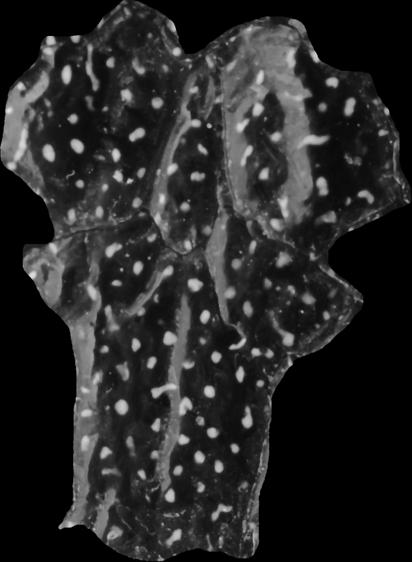}         
        \caption{•}
        \label{fig:MedianImage}
    \end{subfigure}
    \begin{subfigure}[b]{0.3\textwidth}
        \includegraphics[width=\textwidth]{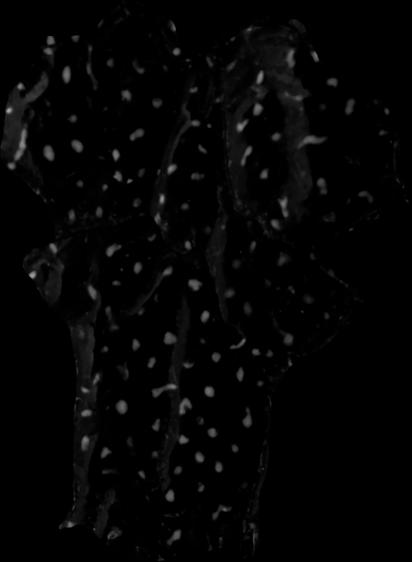} 
        \caption{•}
        \label{fig:GammaImage}
    \end{subfigure}
    \caption{Preprocessed images.}
    \label{fig:ProcessedImages}
\end{figure}

Figure \ref{fig:OriginalImage} shows one original image with spots pattern. Figure \ref{fig:MedianImage} shows an image with gray conversion and median filter. As can be seen, non-homogeneous lighting is not resolved, because it is important to extract the spots in the dark region. Figure \ref{fig:GammaImage} shows an image with the same pre-processing as the Figure \ref{fig:MedianImage} and with a gamma correction, too. Figure \ref{fig:GammaImage} shows that the spots in dark regions are overshadowed. However, the spots in bright regions are not overshadowed, so it is possible to extract the spots that are in bright regions with this pre-processing procedure.

\begin{figure}[h]
\centering
\includegraphics[width=\textwidth]{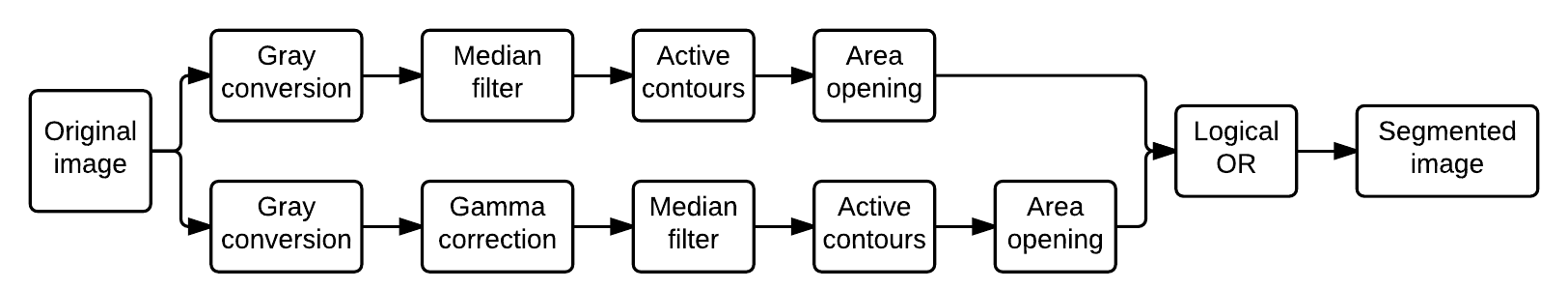}
\caption{Block diagram, segmentation algorithm.}
\label{fig:segmentationAlgorithm}
\end{figure}

The segmentation algorithm is based on the by Giraldo and Salazar proposed algorithm \cite{zuluaga2016automatic}. The segmentation algorithm is divided into two threads. One of the threads consists of four stages; the other one consists of five stages. The stages are gray conversion, median filter, gamma correction, Active Contours and area opening. This procedure is shown in Figure \ref{fig:segmentationAlgorithm}. One thread performs segmentation without gamma correction to extract spots in the dark regions; the other one performs segmentation with gamma correction to extract spots in the bright regions.

The first thread (dark region) consists of a color space transformation that is a linear combination of the original space Red, Green and Blue (RGB) to result in gray space, a median filter to homogenize the region, the principal Active Contour iterations, and finally, the area opening for deleting atypical spots. The image is processed without luminance correction in this thread. The idea of this first thread is to extract spots in the darkest regions.

The second thread (bright region) uses the same color space transformation as the first thread, and a gamma correction procedure. The image is processed with non-linear operation to extract the spots that are in the bright regions in this stage.

The result of Active Contours is a binary image for each thread. Finally, the two threads are merged by a logical or.

\subsection{Ground truth images}

\begin{figure}
    \centering
    \begin{subfigure}[b]{0.4\textwidth}
		\includegraphics[width=\textwidth]{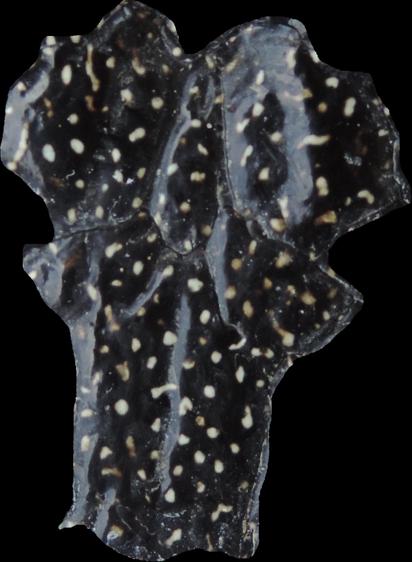}
        \caption{•}
        \label{fig:OriginalImageGT}
    \end{subfigure}
    \begin{subfigure}[b]{0.4\textwidth}
		\includegraphics[width=\textwidth]{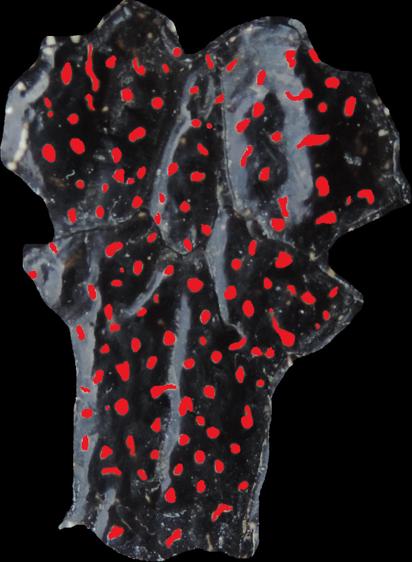}         
        \caption{•}
        \label{fig:GTImage}
    \end{subfigure}
    \caption{Ground truth process.}
    \label{fig:GTImages}
\end{figure}

The ground truth (GT) plays an important role in computer vision to evaluate processes. The GT is important for the development of new algorithms, to compare different algorithms, and to evaluate performance, accuracy and reliability \cite{fernandez2014semi}. For instance, in this paper, Figure \ref{fig:OriginalImageGT} shows one original image of the database and Figure \ref{fig:GTImage} shows its corresponding ground truth image.

\subsection{Iterative Closest Point algorithm}

The Iterative Closest Point algorithm (ICP) was proposed by Besl and Mckay \cite{best1992method} with the idea of resolving registration problems in three dimension $\mathbb{R}^3$. The basic idea is to minimize the difference between two point clouds. The ICP has been used for reconstruction in two and three dimensions, robots localization, bone co-registering, and so forth.

Given two point clousd, the first one A, named source, and the second one B, named target or reference, it desired to minimize the objective function shown in Equation \ref{eqn:objFcnICP}, where $\textbf{R}$ is the rotation matrix and $\textbf{t}$ is the translation vector.

\begin{equation}
Minimize(\sum_{i=1}^{N_a} \left\Vert (\textbf{Ra}_i + \textbf{t}) - \textbf{b}_j \right\Vert^2_2)
\label{eqn:objFcnICP}
\end{equation}

Essentially, the algorithm steps are:

\begin{enumerate}

\item For each point in the source point cloud, find the closest point in the target point cloud.

\item Estimate the combination of rotation and translation using a mean squared error cost function that will best align each source point to its match found in the previous step.

\item Transform the source points using the obtained transformation in the previous step.

\item Iterate until stop criterion, difference between two consecutive values of the objective function, or a maximum number of iterations.

\end{enumerate}

\subsection{Matching algorithm using ICP}

\begin{figure}[h]
\centering
\includegraphics[scale=0.7]{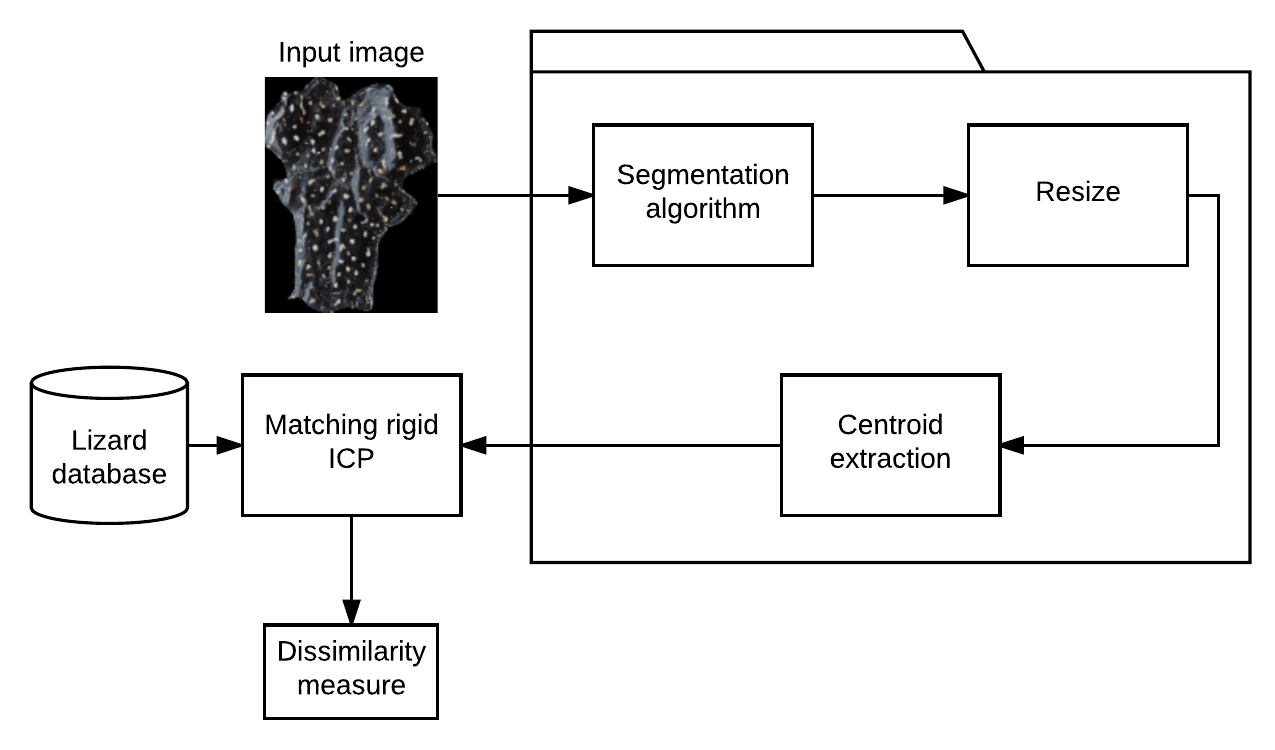}
\caption{Block diagram, matching algorithm 1.}
\label{fig:matAlgICP}
\end{figure}

One of the identification procedures of this paper is based on ICP algorithm. As matching, it has a database to compare with 54 subjects. Figure \ref{fig:matAlgICP} shows the block diagram of the matching algorithm based on ICP. The input image is the segmented image from the segmentation step explained previously, or a ground truth image. This segmented image is re-sized to the same dimension as the scale subject in the database. The centroid of each spot is extracted from the images, and finally the ICP algorithm is executed comparing the input image with each image in the database. The points cloud source are the centroids of the re-sized segmented or ground truth image, and the points cloud target are the centroids of the segmented or ground truth image in the database. Finally, the objective function value is calculated. The procedure explained in the block diagram in Figure \ref{fig:matAlgICP} is executed with all scales in the database. Finally, the subject scale most likely is chosen according to the objective function value in the ICP algorithm.

\subsection{Procrustes analysis}

Procrustes analysis is an optimization technique for spot-pattern matching. The aim of the method is to match the shape of objects generated by the landmark data in the geometric space, and it has advantages in minimizing errors between the landmark data (image in database) and the compared shape (segmented image spots). Procrustes analysis should be equal to compare row (n) of each matrix, such as the reference feature \textbf{X} matrix (nxp) and the comparison feature \textbf{Y} matrix (nxp). It can obtain the rotation error, the translation error, and the scaling error by matching the two data. These errors reflect the comparison of data, and it can make the shape of maximum matching and the reference data \citep{kim2014study}.

\subsection{Matching algorithm using ICP and Procrustes analysis}

\begin{figure}[h]
\centering
\includegraphics[scale=0.7]{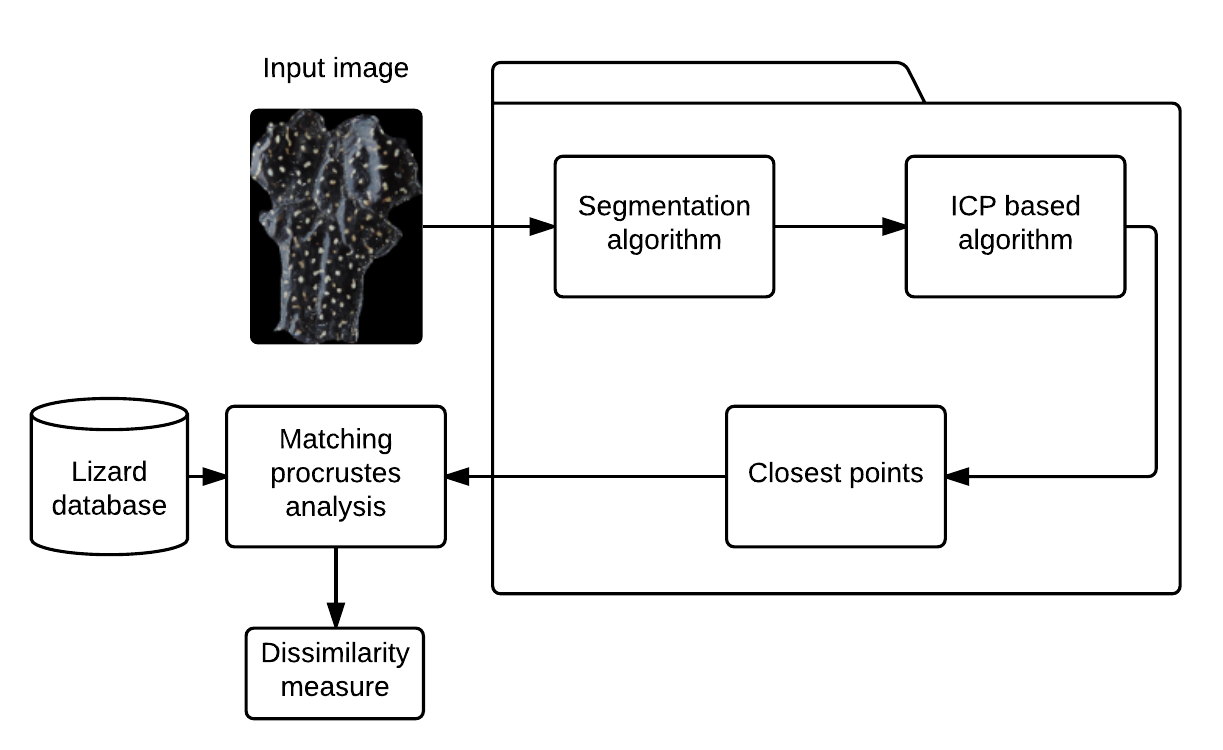}
\caption{Block diagram, matching algorithm 2.}
\label{fig:matAlgProcrustes}
\end{figure}

The other identification algorithm is based on the ICP algorithm and the Procrustes analysis. Figure \ref{fig:matAlgProcrustes} shows the block diagram of this algorithm. This matching algorithm is an extension of the matching algorithm using ICP. After rigid ICP, an algorithm searches the closest points of the base image with reference to the segmented (or ground truth) image, this search is made one to one, i.e. one point in the base image corresponds to one unique point in the segmented (or ground truth) image. Finally, the Procrustes analysis is executed. With the dissimilarity value the most likely or the five most likely scales are found.

The next four subsections are related to the performance metrics to objectively evaluate the segmentation and matching algorithms.

\subsection{Confusion matrix}\label{sec:confMatrix}

Let $\textbf{X}$ be the 2x2 confusion matrix (background and foreground) between the ground truth image and the segmented image, $\textbf{X}_{11}$ is the percentage of the background that was segmented as background, $\textbf{X}_{12}$ is the percentage of the foreground that was segmented as background, $\textbf{X}_{21}$ is the percentage of the background that was segmented as foreground and $\textbf{X}_{22}$ is the percentage of the foreground that was segmented as foreground.

\subsection{Precision, recall and f-measure}

The precision, recall and f-measure are metrics originally applied to machine learning. The value of each metric are extracted with the confusion matrix explained in Section \ref{sec:confMatrix}. Recall is the proportion of the real positive cases that are correctly predicted. Conversely, precision denotes the proportion of predicted positive cases that are correctly real positives. F-measure is the harmonic mean of recall an precision, the metric f-measure gives us an idea of the accuracy of the test \citep{powers2011evaluation}.

\subsection{Hoover metrics}

Hoover metrics \citep{hoover1996experimental} consider five types of regions in the ground truth and machine segmented image comparison, either classified as correctly detected, over-segmented, under-segmented, missed and noise, and then plots the number of areas in each class weighted by total amount of areas based on a threshold (tolerance \%) term that is the free term in which the graphics are based.

\subsection{Performance characteristics}

A biometric identification system, basically is a translation from images to identities. Assuming authentication, biometric systems are required to categorize an individual as genuine (accept) or impostor (reject). For a set of test images presented to the biometric system, the above mentioned classification scheme translates into two error metrics, False Acceptance Rate (FAR) and False Rejection Rate (FRR). The Genuine Acceptance Rate (GAR) is defined as 1-FRR and the Genuine Rejection Rate (GRR) as 1-FAR. The Equal Error Rate (EER) is defined as the cross-over value where FAR and FRR coincide. The EER is often used to compare the performance of different biometric systems \citep{bhattacharyya2009biometric}.

\section{Experimental framework}

This section explains the database conformation and describes the executed experiments. First, the optimization algorithm was executed, followed by the segmentation algorithm and, finally the identification algorithms.

\subsection{Database}

The database consists of 54 individuals (lizards), each one has three frontal scales (samples) in average, for a total of 162 images of frontal scales. The images were taken under controlled conditions. The frontal scales were used to apply the optimization, segmentation, and identification algorithms. The database contains the ground truth and machine segmented images. The database is limited due to the fact that the access to Malpelo Island is restricted at the moment.

\begin{figure}
    \centering
    \begin{subfigure}[b]{0.3\textwidth}
		\includegraphics[width=\textwidth]{1.jpg}
        \caption{Normal}
        \label{fig:normalLigth}
    \end{subfigure}
    \begin{subfigure}[b]{0.3\textwidth}
		\includegraphics[width=\textwidth]{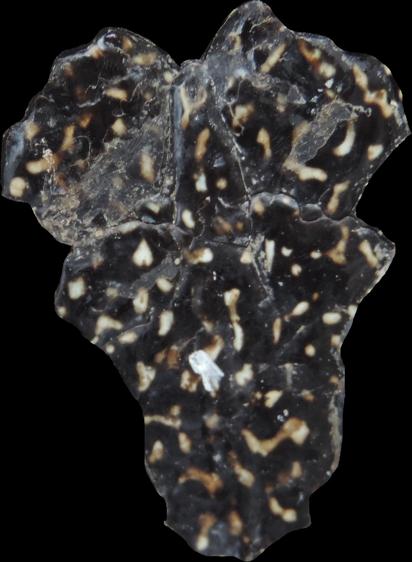}         
        \caption{Ideal}
        \label{fig:idealLigth}
    \end{subfigure}
    \begin{subfigure}[b]{0.3\textwidth}
        \includegraphics[width=\textwidth]{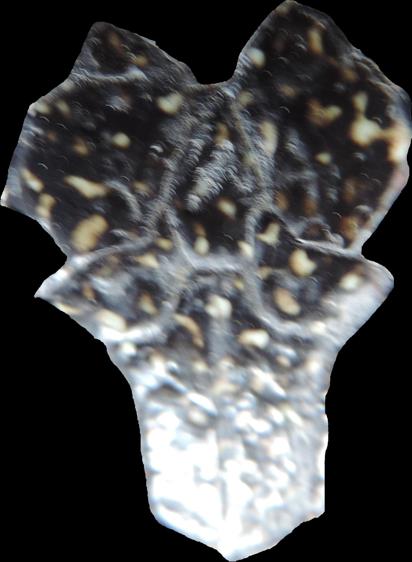} 
        \caption{Hard Exposed}
        \label{fig:hardLigth}
    \end{subfigure}
    \caption{Different light conditions on lizard scales.}
    \label{fig:ligthConditions}
\end{figure}

There are three kinds of light exposition lizard scales: normal (images with normal conditions of luminance), ideal (images without luminance exposure), and hard exposed images (overexposed images). Figure \ref{fig:ligthConditions} shows each kind of light condition that is present in the database.
 	 	 
\subsection{Segmentation experiments}

The parameters of the segmentation algorithm were found by an optimization algorithm \cite{zuluaga2016automatic}. For each segmented image, the binary confusion matrix was extracted with respect to the ground truth. The mean and standard deviation were calculated for all obtained confusion matrices. The precision, recall and f-measure metrics were extracted with respect to the confusion matrix, and the mean and standard deviation were calculated for this metrics. Finally, the Hoover metrics were extracted based on each image in the database and the mean was extracted for visualization purposes.

\subsection{Identification experiments}

To validate the matching algorithms each image in the database was matched against the other ones. From each one of this matches a dissimilarity measure is extracted and saved on a dissimilarity matrix to compute the performance of the algorithm. In this paper two metrics were used to test the performance. The EER metric, this metric was obtained based on the dissimilarity measure of the identification algorithms, and the N-rank, based on the belonging probability. This metric was extracted based on the value of the objective function of registration algorithms. The EER metric was found with the FAR and FRR moving a threshold from 0 to the maximum limit of the dissimilarity matrix. There are three scales of each individual, the N-rank was extracted sorting out the dissimilarities on a descendant order and observing, if the most likely scale belong to the right scale on the three samples by individual for the Top 1, meanwhile on the Top 5 the first, second, third, fourth and fifth likely scales are observed.

The identification algorithms were tested with two versions of the database, the ground truth images database and the automatic machine segmented database. To see the influence of the automatic process on the lizard biometry, the following experiments were carried out. The first experiment takes the ground truth database as target and source, named as ground truth - ground truth (GT-GT). The second experiment takes the automatic database as target and source, named as automatic - automatic (AT-AT). The final experiment takes the ground truth database as target and the automatic segmented database as source, named as ground truth - automatic (GT-AT).

\section{Results and Discussion}

This section shows the experimental results and discussion of the segmentation and identification algorithms.

\subsection{Segmentation results}\label{sec:segmeResults}

Table \ref{tbl:confMatrix} shows the result of the segmentation algorithm. The performance of the segmentation algorithm is 82.87\% that is $(X_{11}+X_{22})/2$, see the Section \ref{sec:confMatrix}. 

\begin{table}[h]
\centering
\caption{Confusion matrix results related to the segmentation experiments.}
\label{tbl:confMatrix}
\begin{tabular}{|c|c|c|}
\hline
           & Background        & Foreground         \\ \hline
Background & 96.21\% $\pm$ 1.98\% & 30.48\% $\pm$ 25.55\% \\ \hline
Foreground & 3.79\% $\pm$ 1.98\%  & 69.52\% $\pm$ 25.55\% \\ \hline
\end{tabular}
\end{table}

Table \ref{tbl:clasMetrics} shows the classical metric precision, recall and f-measure. Each classical metric was extracted with the images of the validation set, and the mean and standard deviation are shown.

\begin{table}[h]
\centering
\caption{Classic metrics for the segmentation algorithm validation.}
\label{tbl:clasMetrics}
\begin{tabular}{|c|c|c|c|}
\hline
Metric & Precision          & Recall             & F-measure          \\ \hline
Value  & 70.33\% $\pm$ 24.57\% & 40.13\% $\pm$ 16.58\% & 48.44\% $\pm$ 17.91\% \\ \hline
\end{tabular}
\end{table}

The metrics in Table \ref{tbl:clasMetrics} show that 40.13\% of the foreground is correctly predicted by the model. The general performance of the automatic segmentation procedure is 48.44\% given by the F-measure. The precision metric gives us a notion of correct background over all supposed background in the segmentation algorithm with a performance of 70.33\%, this means that the presented segmentation method has a better prediction for the background regions than the foreground regions. There is an unavoidable human error on the ground truth due to the difficulty of segmenting the borders. The results of the segmentation algorithm in the validation have high standard deviation results, the classical metrics results support this affirmation, where the standard deviation of each metric is close to 20\%, as a consequence of the variability of the light exposure in the database. This means that the pre-processing step should be improved.

\begin{figure}[hp]
    \centering
    \begin{subfigure}[b]{0.4\textwidth}
		\includegraphics[width=\textwidth]{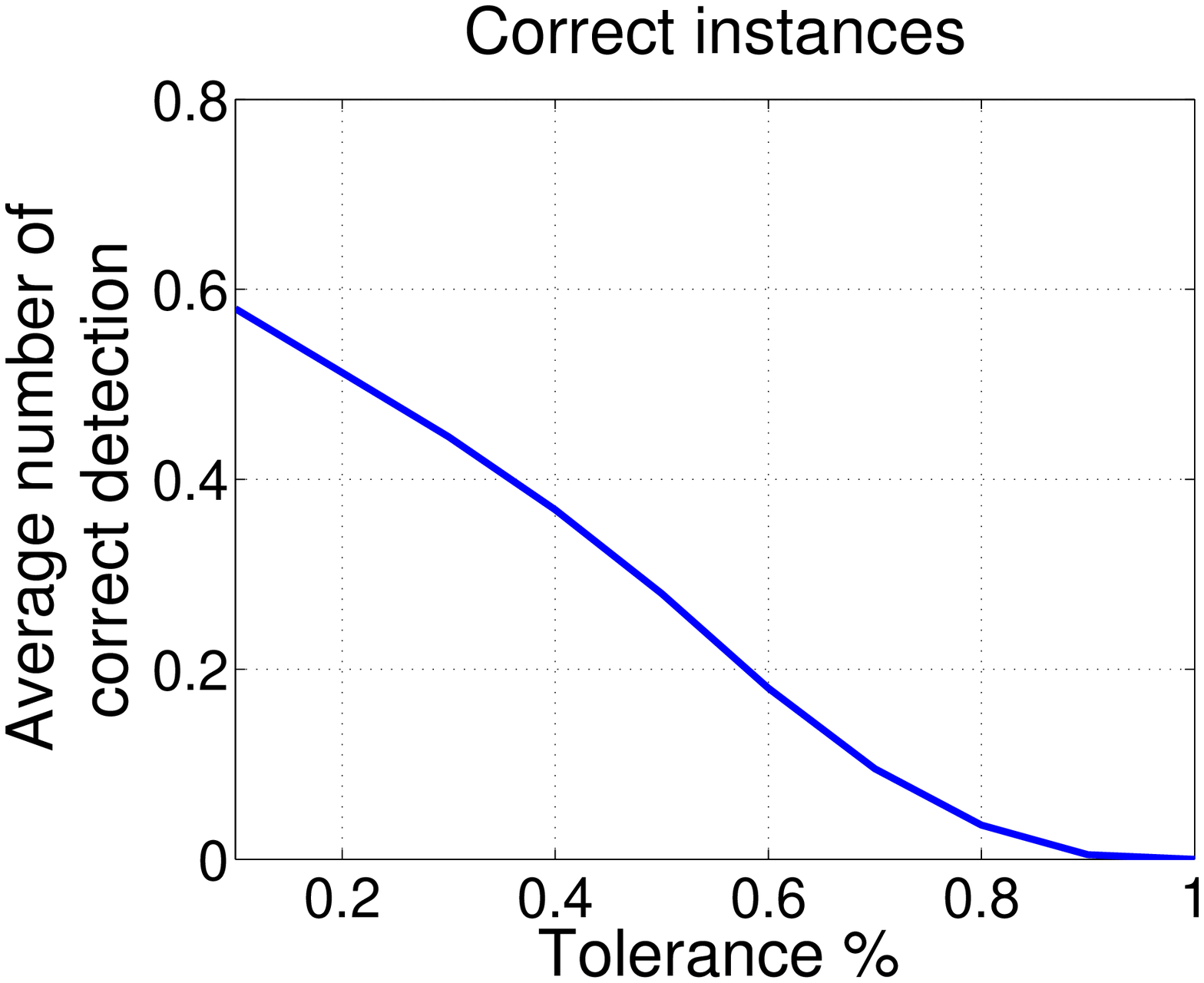}
        \caption{•}
        \label{fig:corrDetec}
    \end{subfigure}
    \begin{subfigure}[b]{0.4\textwidth}
		\includegraphics[width=\textwidth]{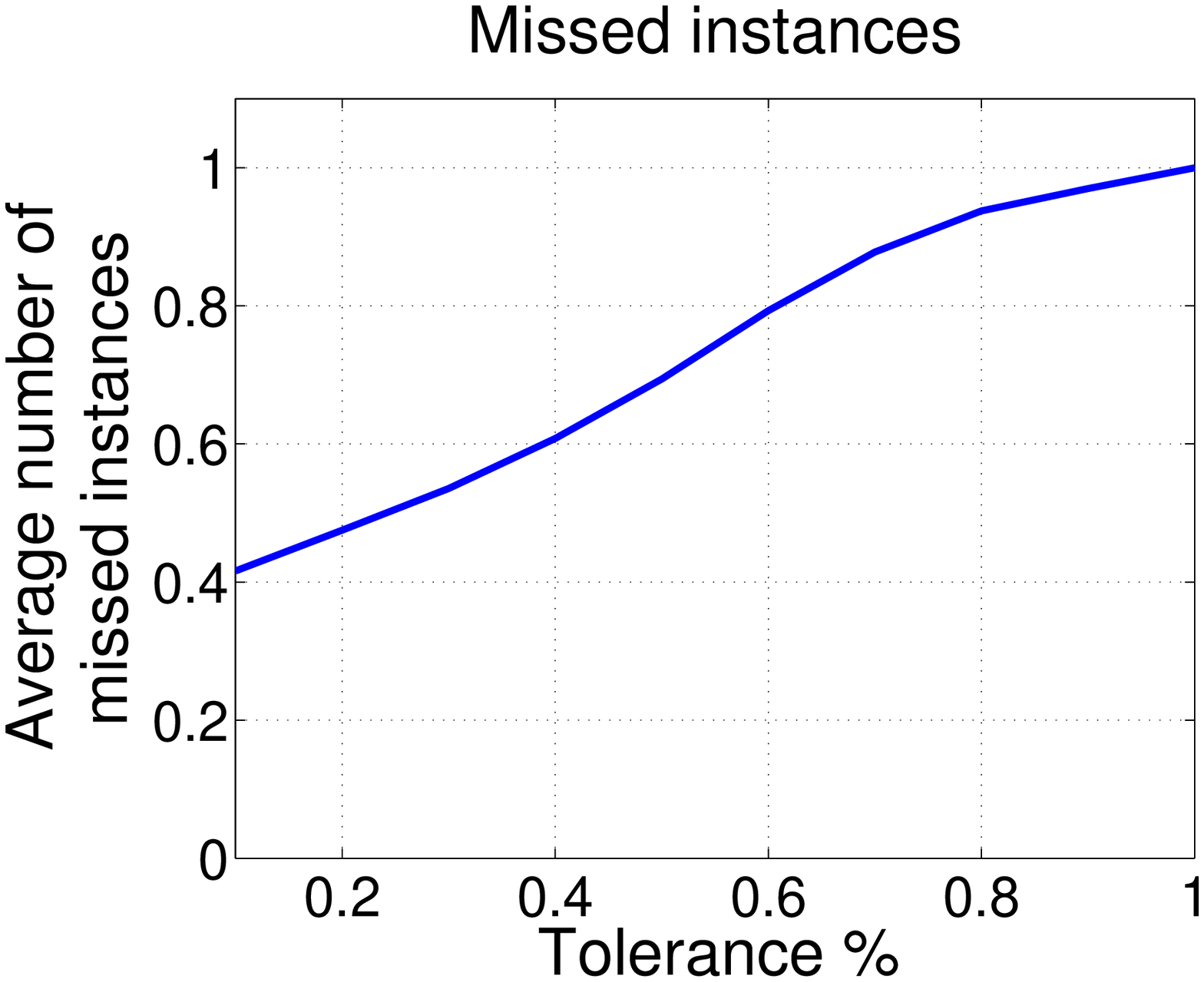} 
        \caption{•}
        \label{fig:missInst}
    \end{subfigure}
    \begin{subfigure}[b]{0.4\textwidth}
		\includegraphics[width=\textwidth]{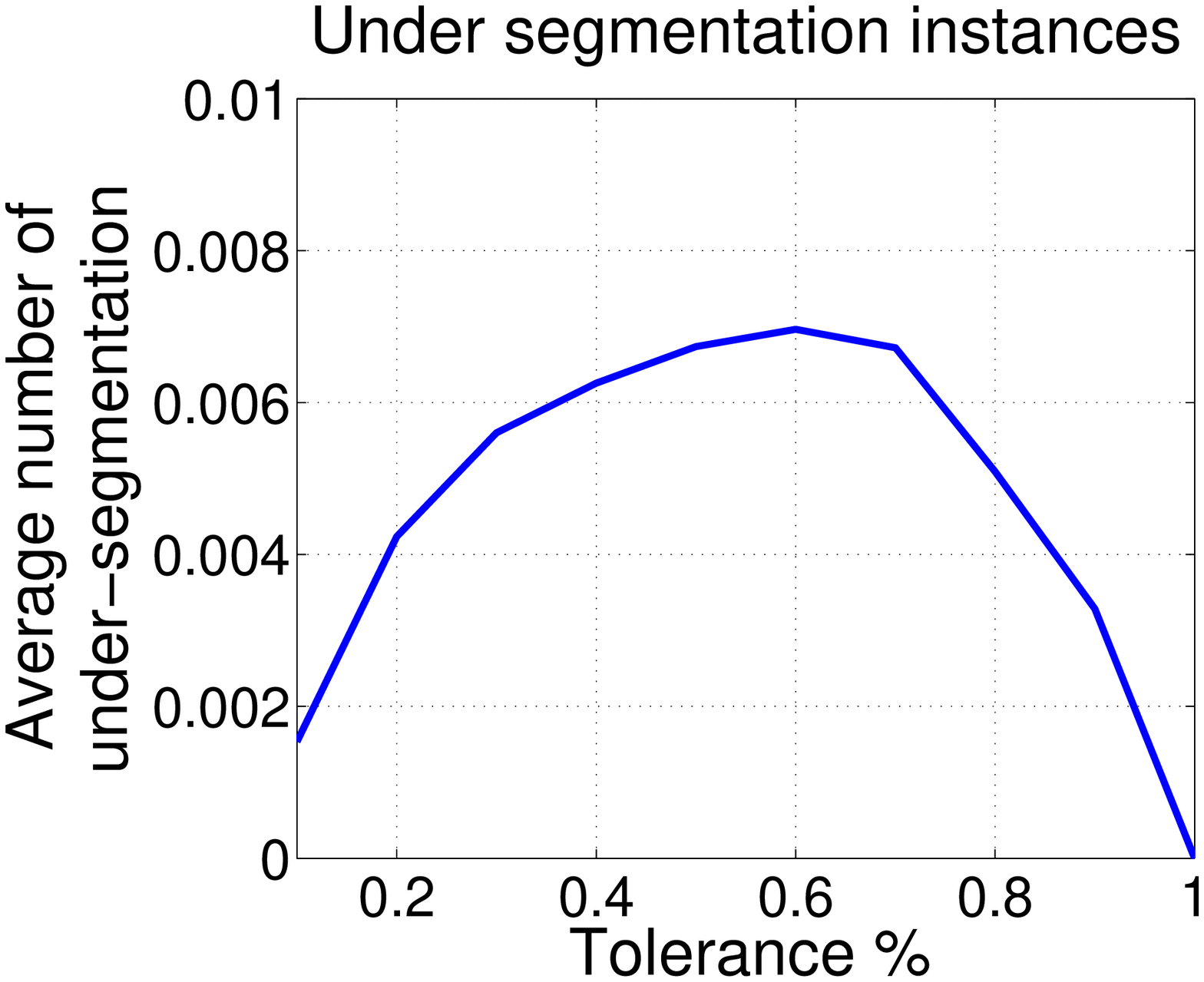}
        \caption{•}
        \label{fig:underSegme}
    \end{subfigure}
    \begin{subfigure}[b]{0.4\textwidth}
		\includegraphics[width=\textwidth]{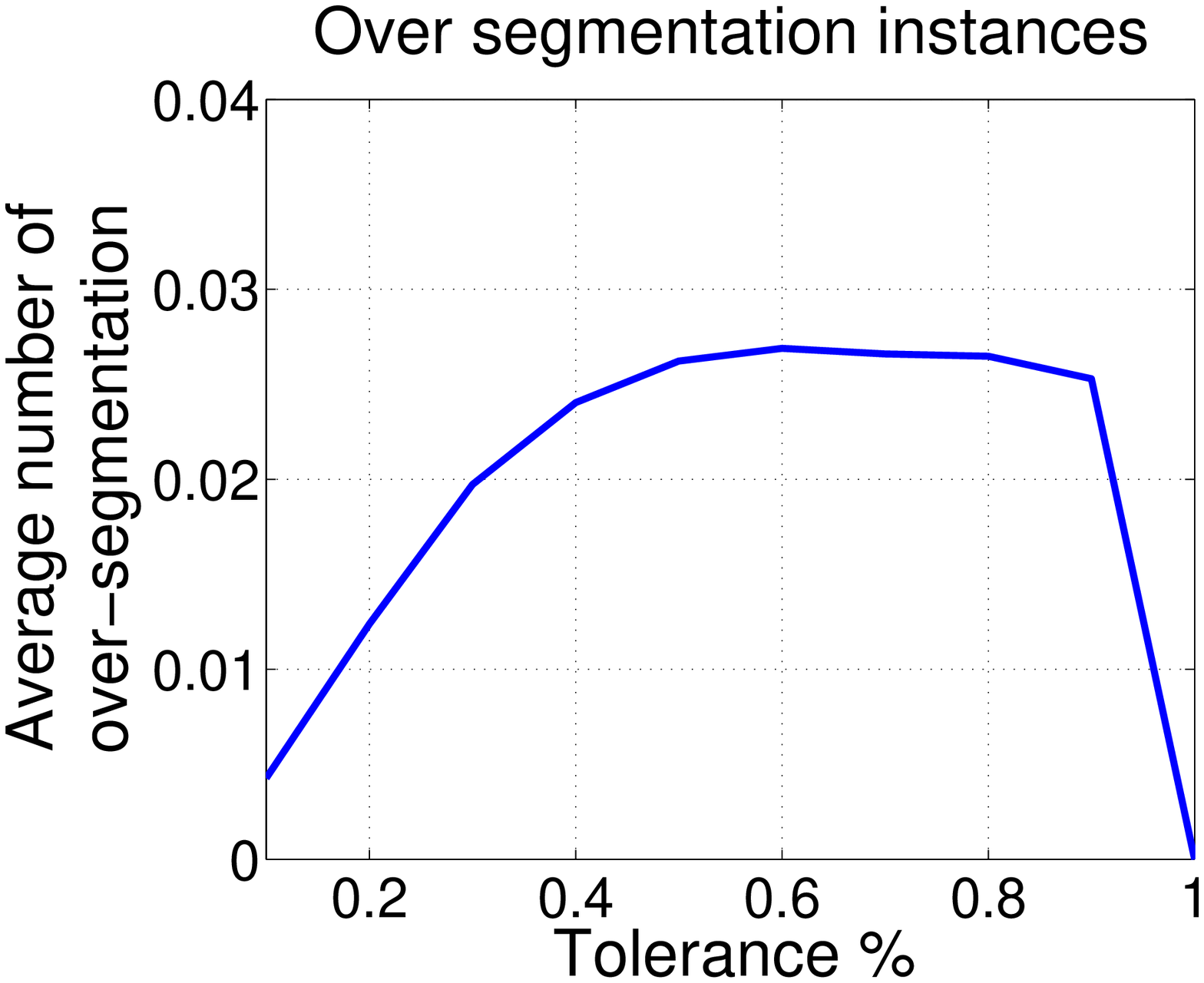} 
        \caption{•}
        \label{fig:overSegme}
    \end{subfigure}
    \begin{subfigure}[b]{0.4\textwidth}
        \includegraphics[width=\textwidth]{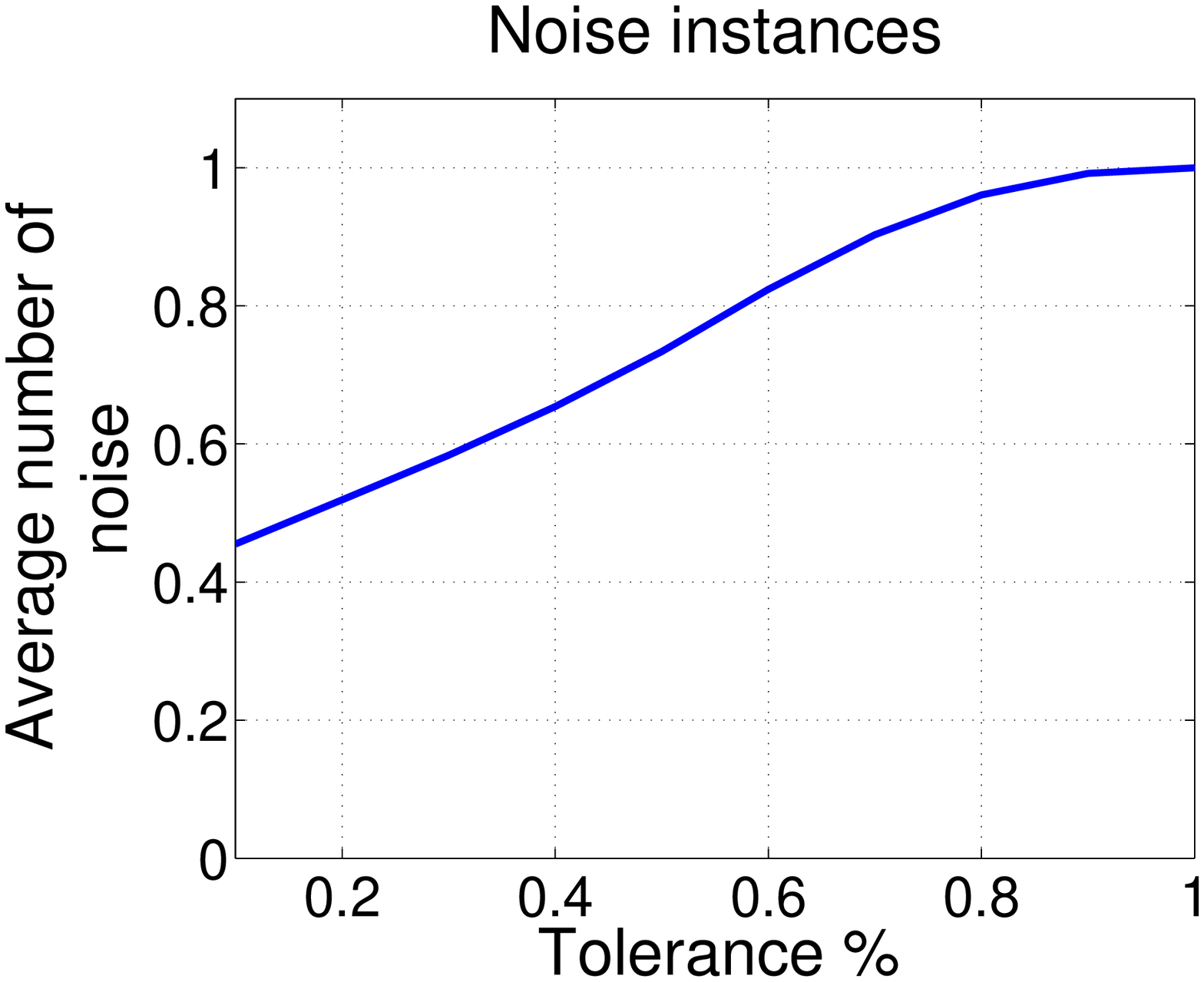}
        \caption{•}
        \label{fig:noise}
    \end{subfigure}
    \caption{Average of Hoover metrics related to the segmentation algorithm.}
    \label{fig:HooverMetrics}
\end{figure}

Figure \ref{fig:HooverMetrics} shows the average Hoover metrics of the entire database. Each figure was extracted with the automatic segmented and ground truth images, and then the mean of each metric was computed.

Figure \ref{fig:corrDetec} shows the average number of correct instances under a percentage of tolerance. Figure \ref{fig:corrDetec} shows 3.59\% of average performance using active contours with 80\% of tolerance, i.e. with a certain tolerance value there is an average amount of images correctly segmented. There is a 0\% performance with 100\% of tolerance, due to the fact that there is not segmented images that are totally equal to some ground truth images. A 100\% performance with 100\% tolerance is practically impossible due to the borders problem on the images format, or human errors on the manually segmented images. Figure \ref{fig:missInst} shows another point of view of the correct instances. Figure \ref{fig:missInst} shows 93.76\% of missed instances with 80\% of tolerance.

Figure \ref{fig:underSegme} demonstrates that there are no under segmentation problems due to the active contour approach. Figure \ref{fig:overSegme} shows that over segmentation problems are minimal. Figure \ref{fig:noise} shows 70\% of noise in average. Figures \ref{fig:missInst}, \ref{fig:overSegme} and \ref{fig:noise} give us an indication that the biggest problem is the high-exposed scales, because the little luminous regions are classified by the active contour technique as a spot. Meanwhile the large bright regions are erased, even with the spots that are inside this regions, by the area opening algorithm.

The average running time of the segmentation algorithm was 143.9 $\pm$ 53.3 seconds per each segmented image. The experiments were carried out on an Intel Core i7 3630QM with 8 gigabytes of RAM memory.

\subsection{Identification results}

\begin{figure}[hp]
    \centering
    \begin{subfigure}[b]{0.4\textwidth}
		\includegraphics[width=\textwidth]{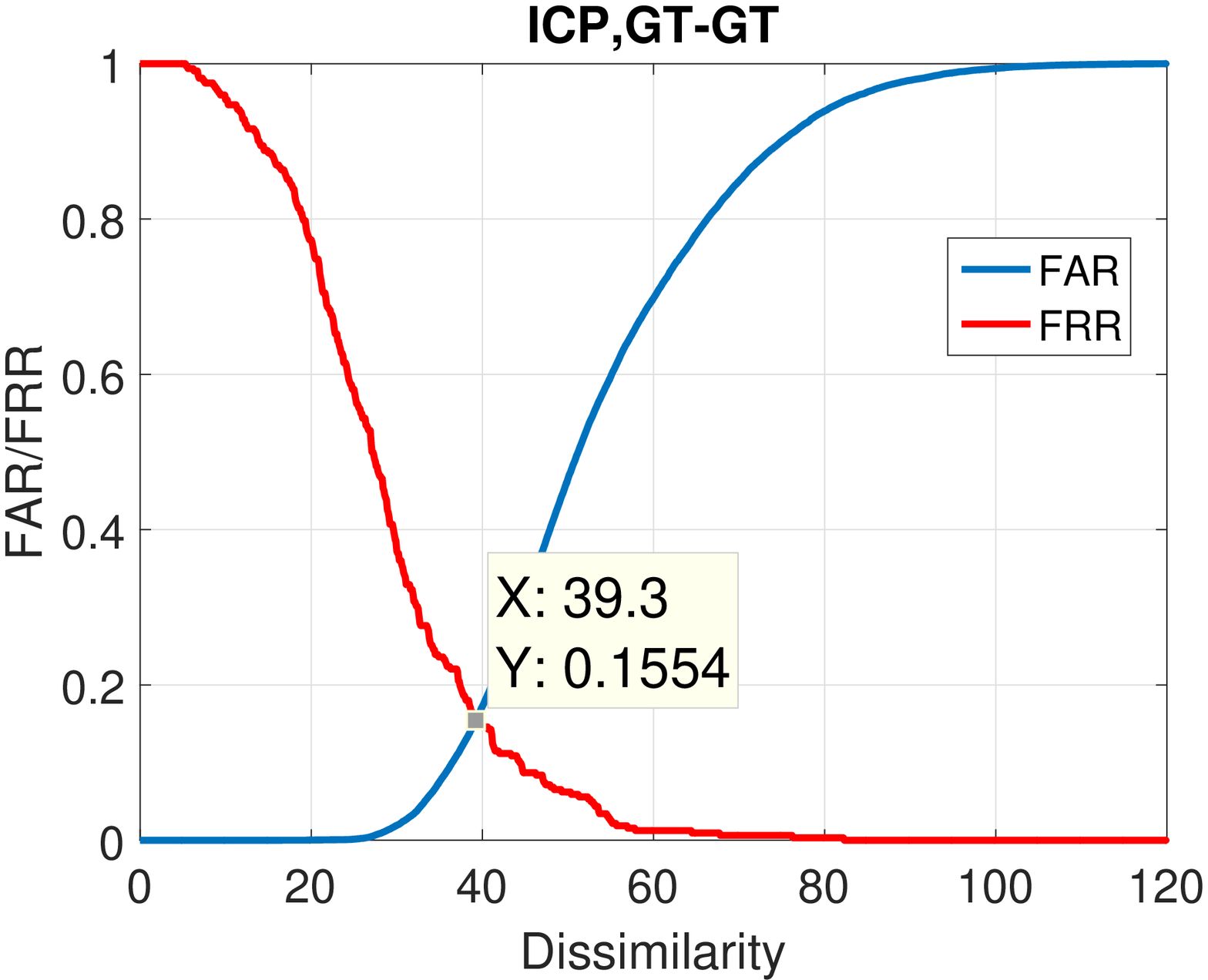} 
        \caption{•}
        \label{fig:ICPGT-GT}
    \end{subfigure}
    \begin{subfigure}[b]{0.4\textwidth}
        \includegraphics[width=\textwidth]{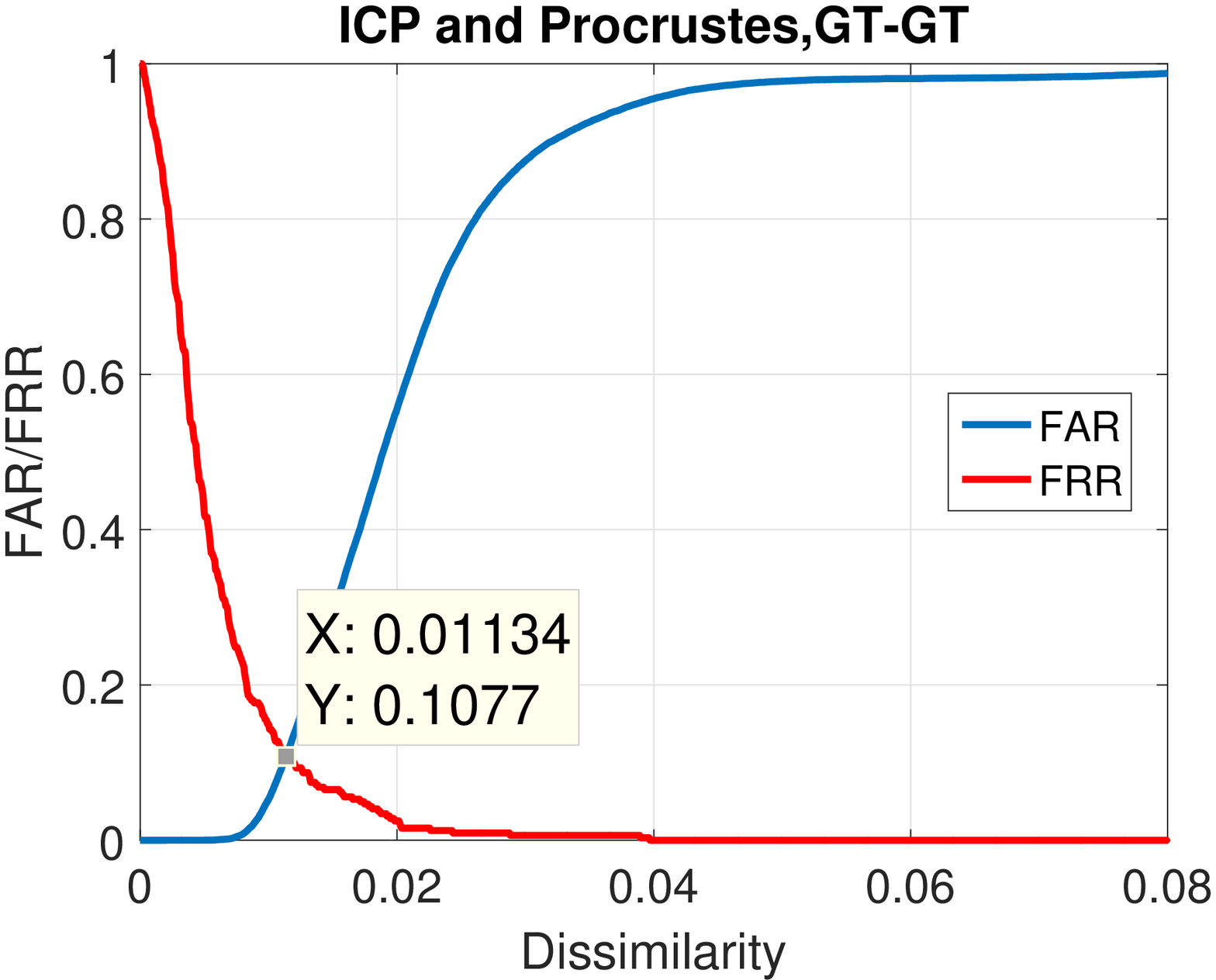} 
        \caption{•}
        \label{fig:ICPProGT-GT}
    \end{subfigure}
    \begin{subfigure}[b]{0.4\textwidth}
		\includegraphics[width=\textwidth]{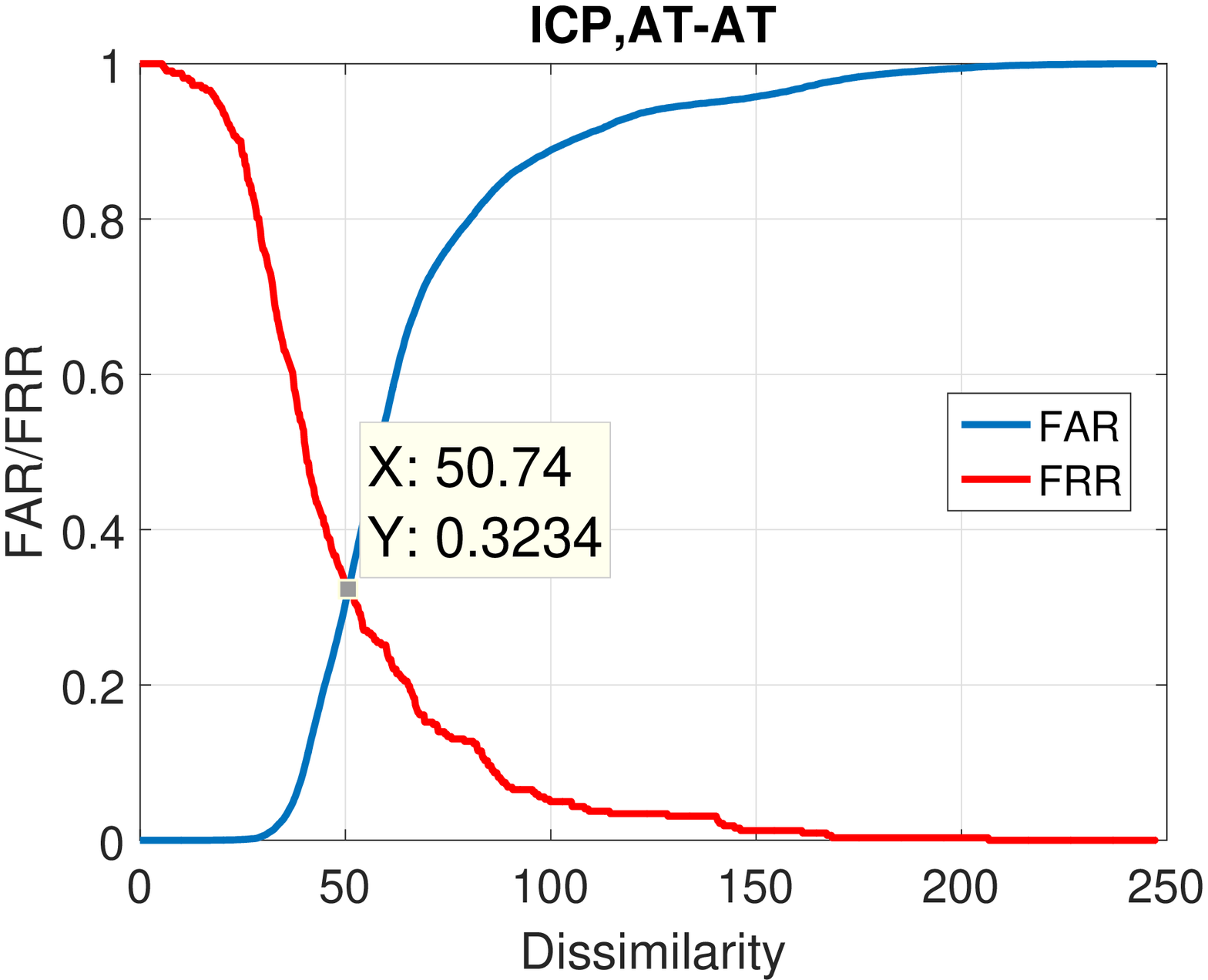} 
        \caption{•}
        \label{fig:ICPAT-AT}
    \end{subfigure}
    \begin{subfigure}[b]{0.4\textwidth}
		\includegraphics[width=\textwidth]{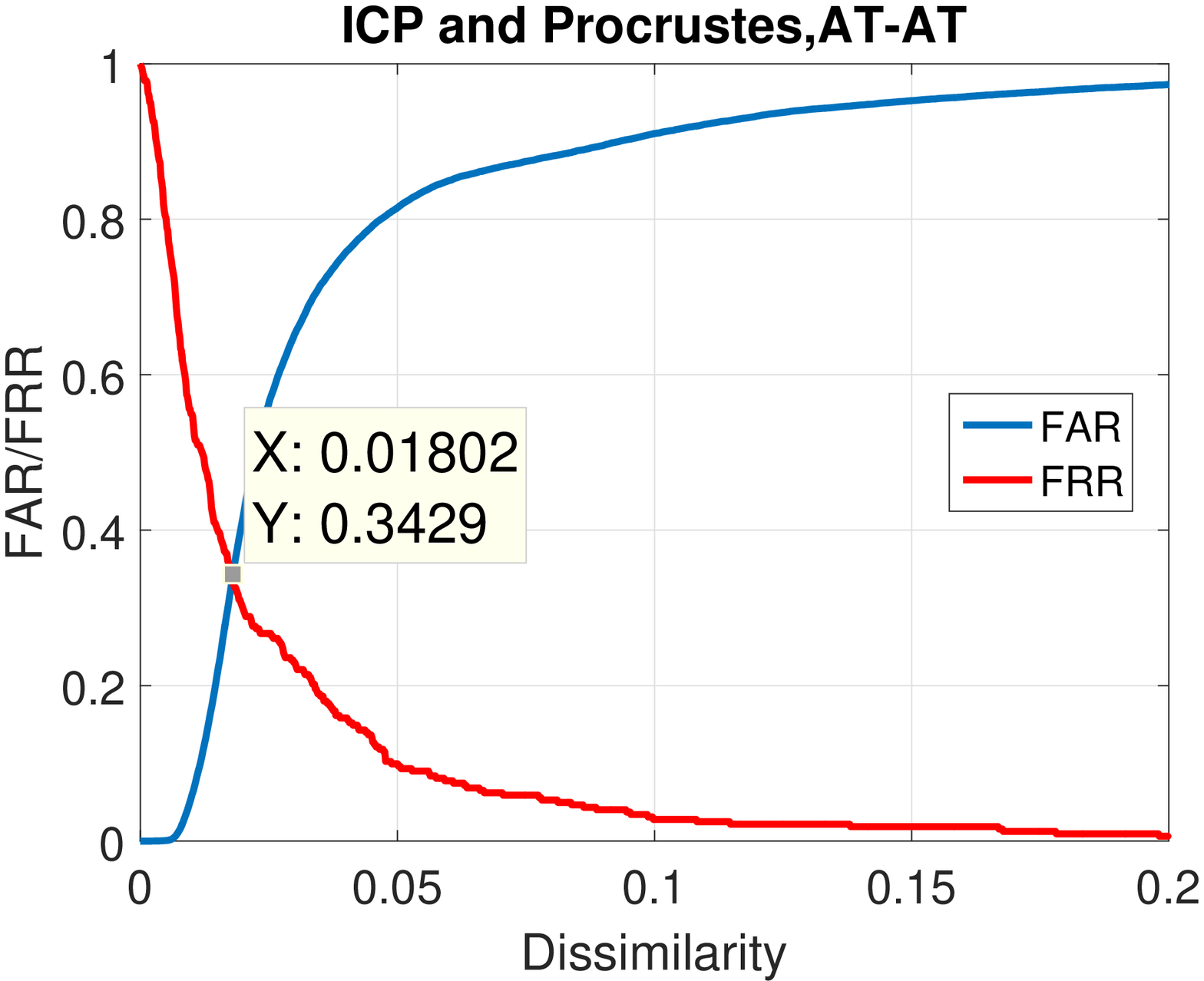} 
        \caption{•}
        \label{fig:ICPProAT-AT}
    \end{subfigure}
    \begin{subfigure}[b]{0.4\textwidth}
        \includegraphics[width=\textwidth]{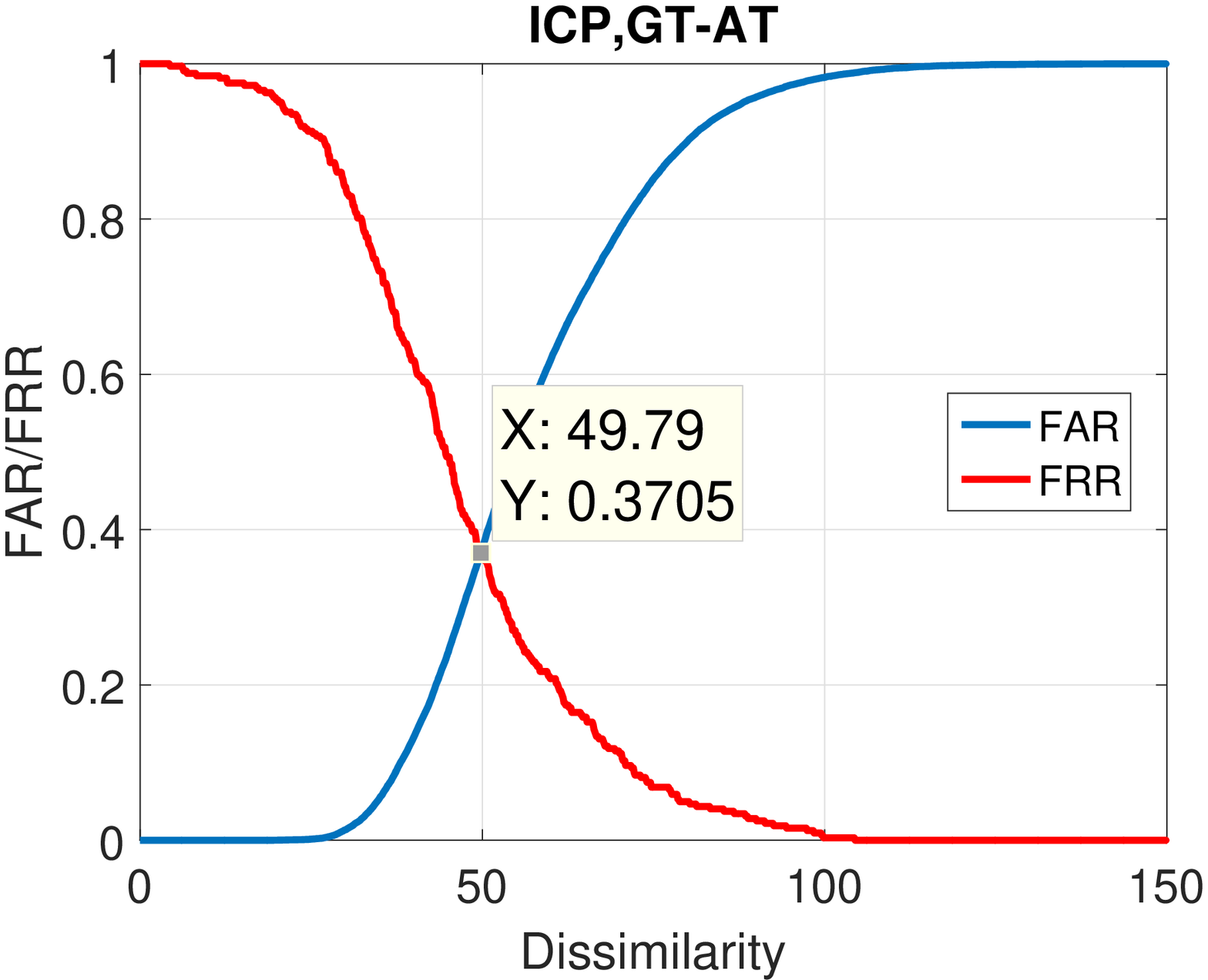}  
        \caption{•}
        \label{fig:ICPGT-AT}
    \end{subfigure}
    \begin{subfigure}[b]{0.4\textwidth}
        \includegraphics[width=\textwidth]{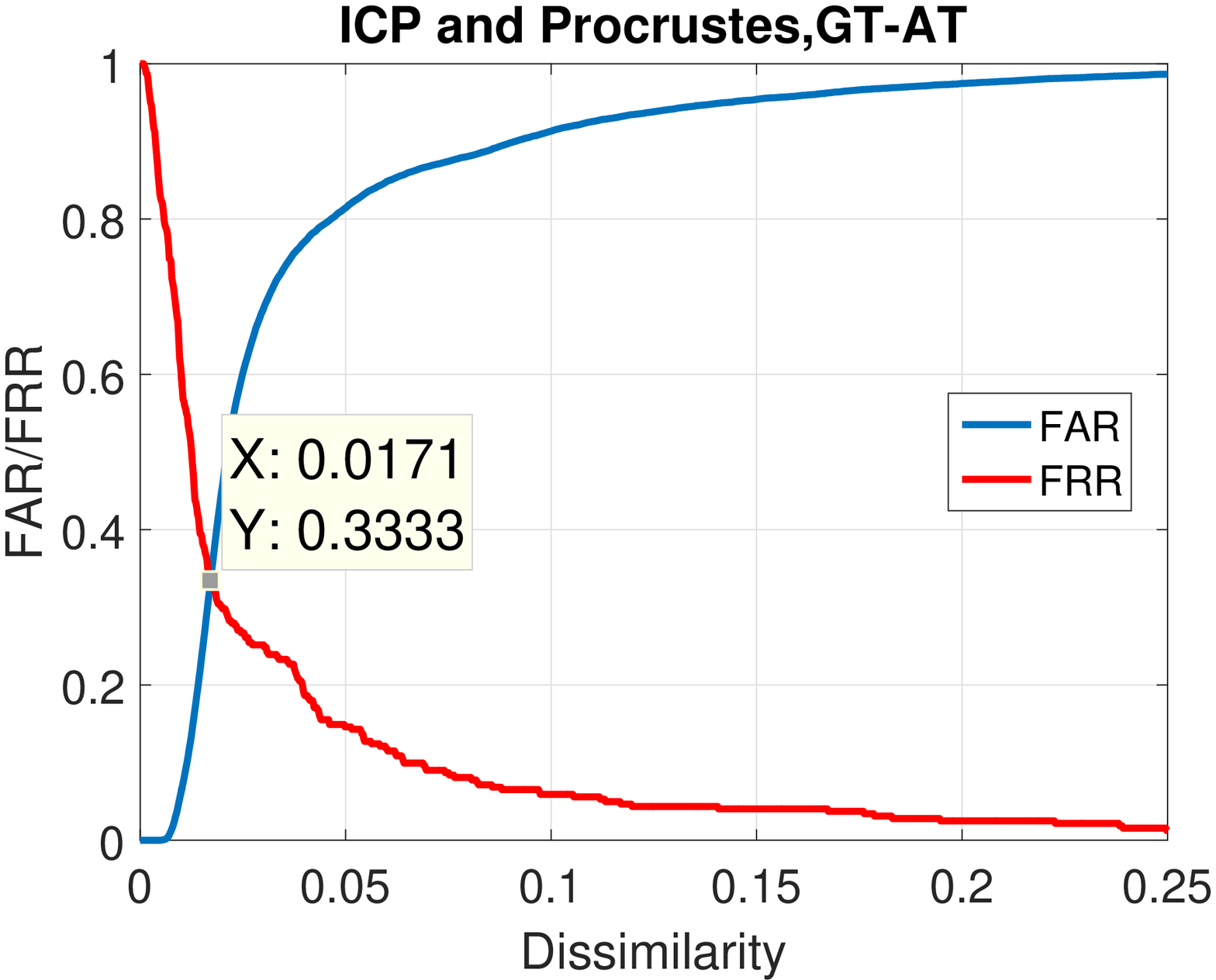} 
        \caption{•}
        \label{fig:ICPProGT-AT}
    \end{subfigure}
    \caption{FAR and FRR functions for identification experiments.}
    \label{fig:EER}
\end{figure}

Figure \ref{fig:EER} shows the FAR and FRR functions depending on the threshold, and based on this the EER was computed as the cross-over value of the FAR and FRR function. Table \ref{tbl:EERResults} shows the summary of EER results. The EER metrics were extracted with respect to the dissimilarity results matrix. These metrics were computed moving a threshold between 0 and the maximun number of the dissimilarity matrix. The threshold movement computed the FAR and FRR functions. Those functions were summarized in Figure \ref{fig:EER} for each experiment. The interception point between the FAR and the FRR functions was marked for the explicit visualization.

Tables \ref{tbl:EERResults} and \ref{tbl:perforBelongProb} show the results of the two identification algorithms that were tested. With the Equal Error Rate, one optimal threshold based on the dissimilarity is chosen. Meanwhile the N-rank metric based on the belonging probability chooses the most similar individual to the input scale for the Top 1, and the five most similar individuals to the input scale for the Top 5. The advantage of the EER metric is that if one individual is not matched with the database, there is a probability of 1-EER that the scale belongs to a new individual.

There is an 89.23\% probability that the described algorithm chooses a new individual correctly. Table \ref{tbl:EERResults} shows a minimum EER of 10.77\% on the GT-GT experiments with the ICP and Procrustes matching algorithm.

\begin{table}[h]
\centering
\caption{ERR of the identification algorithms.}
\label{tbl:EERResults}
\begin{tabular}{|c|c|c|}
\hline
      & ICP (EER) & ICP \& Procrustes (EER) \\ \hline
GT-GT & 15.54\%   & 10.77\%                   \\ \hline
AT-AT & 32.34\%   & 34.29\%                  \\ \hline
GT-AT & 37.05\%   & 33.33\%                  \\ \hline
\end{tabular}
\end{table}

There is a gap error and performance between the GT-GT experiments and AT-AT, GT-AT experiments in both metrics. The differences between the GT-GT and the other experiments are due to the contrast among ground truth and automatic segmented images discussed in Section \ref{sec:segmeResults}. It is important to improve the segmentation algorithm to enhance the identification results in order to design an automatic lizard biometric system.

\begin{table}[h]
\centering
\caption{N-rank metric related to identification algorithms.}
\label{tbl:perforBelongProb}
\begin{tabular}{|c|c|c|c|c|}
\hline
Matching Algorithm & \multicolumn{2}{c|}{ICP} & \multicolumn{2}{c|}{ICP \& Procrustes} \\ \hline
Top                & Top 1       & Top 5      & Top 1              & Top 5              \\ \hline
GT-GT              & 69.43\%     & 83.44\%    & 92.99\%            & 96.82\%        \\ \hline
AT-AT              & 40.76\%     & 57.32\%    & 60.51\%            & 77.71\%        \\ \hline
GT-AT              & 18.47\%     & 37.58\%    & 57.32\%            &    75.80\%        \\ \hline
\end{tabular}
\end{table}

Table \ref{tbl:perforBelongProb} shows a maximum performance of 96.82\% on the Top 5 of the GT-GT experiments with the ICP and Procrustes matching algorithm. The Top 1 gives the correct individual with a performance of 92.99\% if the input individual belongs to the database. The Top 5 gives the five most likely individuals to choose the correct individual with human supervision. The Top 5 has a performance of 96.82\% if the input individual is in the database.

The average running time of the ICP experiment was 12.66 $\pm$ 2.12 seconds for each matching image. The average running time of the ICP and Procrustes experiment was 33.58 $\pm$ 9.12 seconds for each matching image. The implementation exploits the parallelism of the processor, because the algorithm has a practical application. If we augment the core number of the processor the time of the identification algorithm decreases. If new individuals are added to the database the time of the identification algorithm increases. The experiments were carried out on an Intel Core i7 3630QM with 8 gigabytes of RAM memory.

\subsection{Software}

\begin{figure}[h]
\centering
\includegraphics[width=\textwidth]{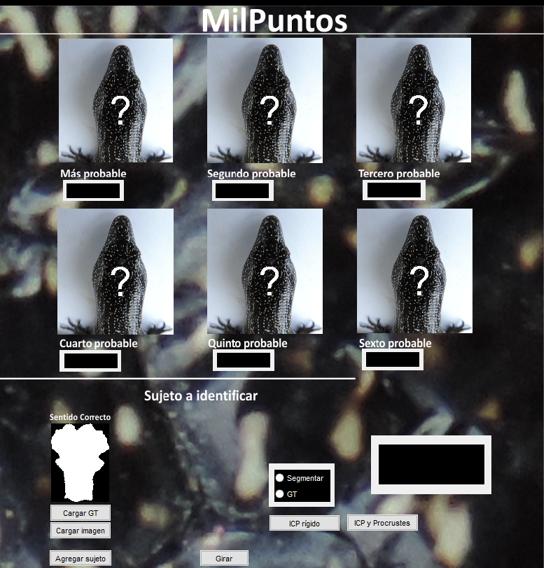}
\caption{Screen-shot of the software MilPuntos.}
\label{fig:screenShot}
\end{figure}

MilPuntos is an identification software to assist in the biological task of \textit{Diploglossus Millepunctatus} lizard species. MilPuntos is a software developed in Matlab with the algorithm introduced in this paper. The software is developed entirely on M language of Matlab. This is the first version of the software. The software has been tested on a computer with an Intel Core i7 3630QM processor, 8 gigabytes RAM on Matlab 2013b. Figure \ref{fig:screenShot} shows a screen-shot of the software MilPuntos. The software includes a local database, the segmentation algorithm, the identification procedure with the belonging probability criterion of selection, and an option to add new individuals to the database.

The images in the local database have the same orientation. The automatic algorithm ensures a vertical orientation for the input image, but the direction should be ensured manually by the software user. Figure \ref{fig:frontalScale} shows the specific scale that should be manually segmented as algorithm input.

The database and the created software is publicly available for download on the project link \cite{Salazar2016lizardDatabase}.

\section{Conclusions and Future work}

We introduced an algorithm for automated segmentation of lizard spots and identification of \textit{Diploglossus Millepunctatus}. The segmentation algorithm is composed of a pre-processing stage, active contours iterations and morphological filtering. The best parameters for the three segmentation algorithm stages were selected using an optimization with the objective to segment true spots and to correct the background. The optimization algorithm needs a better design with respect to practical applications, e.g. with other animals, because the processing time is already quite high. Another important aspect is that the segmentation algorithm execution, after optimization, is not as time-demanding as other segmentation algorithms in the state of the art, e.g. Markov Random Field, graph cuts and so on. The identification algorithm is composed of an normalization stage, centroid point extraction and registration algorithms execution. For the FAR/FRR experiments, the optimal thresholds of dissimilarities named EER were chosen through an exhaustive search. The general performance of the automatic segmentation procedure is 48.44\% given by the F-measure. We reach 92.99\% and 96.82\% of performance on Top 1 and Top 5 respectively for manual-segmented images. We need to improve the segmentation stage, because the best performance for machine-segmented images was 57.32\% in the Top 5. The ICP and Procrustes algorithm present better results than the ICP matching algorithm. The database and the created software is publicly available for download on the project link.

For future work the high exposure problem will be addressed with a better pre-processing technique. It is important to evaluate the change of the spot patterns along the time, with the objective of analyze the robustness of the system in the time, e.g. evaluate the algorithms performance in some consecutive years maping the change of the spot patterns.\\
\\
\textbf{Acknowledgment.} The authors express thanks to Alexander G\'omez Villa and German D\'iez Valencia for their support in the database preparation. Also, Fundaci\'on Malpelo and Parques Nacionales de Colombia provided funding and research permits for collecting data at the Santuario de Fauna y Flora Isla de Malpelo.

\section*{References}

\bibliographystyle{unsrt}
\bibliography{bibfile}

\end{document}